%% file: DEGREE-netV3-sub.tex
\documentclass[runningheads]{llncs}
\usepackage{graphicx}
\usepackage{amsmath,amssymb} 
\usepackage{color}
\usepackage{overpic}
\usepackage[ruled]{algorithm2e}
\usepackage{soul}
\usepackage{xcolor,colortbl}
\usepackage{multirow}
\usepackage{subfigure}

\newcommand{\M}[1]{{\color{black}#1}}
\newcommand{\RM}[1]{{\color{black}#1}}
\newcommand{\RRM}[1]{{\color{black}#1}}
\newcommand{\RRRM}[1]{{\color{black}#1}}
\newcommand{\RRRRM}[1]{{\color{black}#1}}
\newcommand{\RRRRRM}[1]{{\color{black}#1}}
\newcommand{\RsixM}[1]{{\color{black}#1}}
\usepackage[width=122mm,left=12mm,paperwidth=146mm,height=193mm,top=12mm,paperheight=217mm]{geometry}

\definecolor{Gray}{gray}{0.85}

\setstcolor{red}

\def\HR{result}
\def\ANR{A+}
\def\CNN{CNN}
\def\CSCN{CSCN}
\def\JSBNE{JSBNE}
\def\SESR{SelfExSR}
\def\EGRCNN{EGRCNN}

\begin{document}
	\pagestyle{headings}
	\mainmatter
	\def\ECCV14SubNumber{72}
	
	\title{Deep Edge Guided Recurrent Residual Learning for Image Super-Resolution}
	\vspace{-4mm}
	
	\titlerunning{ECCV-16 Submission}
	
	\authorrunning{Wenhan Yang \textit{et al.}}
	
	\author{Wenhan Yang$^{1}$, Jiashi Feng$^{2}$, Jianchao Yang$^{3}$, Fang Zhao$^{2}$, Jiaying Liu$^{1}$, \\
		Zongming Guo$^{1}$, Shuicheng Yan$^{2}$}
	\institute{$^{1}$Peking University, $^{2}$National University of Singapore, $^{3}$Snapchat Inc.}

	\maketitle
	\vspace{-8mm}
	
	\begin{abstract}
	\M{In this work, we consider the  image super-resolution (SR) problem. The main challenge of image SR is to recover \RRM{high-frequency}  details of a low-resolution (LR) image that are important for human perception.
	To  address this essentially ill-posed problem,  we introduce a Deep Edge Guided REcurrent rEsidual~(DEGREE) network to progressively  recover the  high-frequency details. Different from most of existing methods that aim at predicting high-resolution (HR) images directly, DEGREE  \RsixM{investigates} an alternative route to recover the difference between a pair of LR and HR images by recurrent residual learning. 
	\RsixM{DEGREE further augments the SR process with edge-preserving capability, namely the LR image and its edge map can jointly infer the sharp edge details of the HR image during the recurrent recovery process.}
	To speed up its training convergence rate, by-pass connections across multiple layers of DEGREE are constructed. \RsixM{In addition}, we offer an understanding on DEGREE from the view-point of sub-band frequency decomposition on image signal and experimentally demonstrate how DEGREE can  recover different frequency bands separately. Extensive experiments on three benchmark datasets clearly demonstrate the superiority of DEGREE over well-established baselines and DEGREE also provides new state-of-the-arts on these datasets.	
	}
		
	\end{abstract}
	\vspace{-9mm}
	
	\section{Introduction}
	\vspace{-2mm}
	
	Image super-resolution (SR) aims at recovering a high resolution (HR) image from  low resolution (LR) observations. 
	Although \RRM{it} has seen wide applications, \RRM{such as surveillance video recovery~\cite{Zhang2010848}, 
		face hallucination~\cite{Liu2007}, medical image enhancement~\cite{SRMI},}
	the SR problem, or more concretely the involved inverse signal estimation problem therein, is essentially ill-posed and still rather difficult to solve. \RRM{In order to relieve  ill-posedness of the problem, most of  recent SR methods propose to incorporate various {prior} knowledge about natural images to regularize the signal recovery process. 
	This  strategy establishes  a standard maximum \emph{a posteriori} (MAP) image SR framework~\cite{MAP1,MAP2}, where an HR image is  estimated by maximizing its fidelity to the  target  with kinds of \emph{a priors}.}

\RM{
	Most of existing MAP based image SR methods~\cite{yang_image_2010,sun2008image} associate the data fidelity term  
	with the mean squared error (MSE), in order to ensure consistency between the estimated HR image and the ground truth when learning model parameters. However, solely considering minimizing MSE 
	usually fails to recover the sharp or high-frequency details such as textures and edges. This phenomenon is also observed in much previous  literature~\cite{sub-band1,sub-band2,singh2014super,Song-ICASSP-2016}.
	To address this problem, bandpass filters -- that are commonly used to extract texture features -- were employed to preserve sharp details in the image SR process\RRRM{~\cite{sub-band1,Song-ICASSP-2016,sub0,Chatterjee2007}}. The bandpass filters decompose an LR image into several \RRM{sub-band images} and  build hierarchical fidelity terms to steer  recovery of those \RRM{sub-band images}. \RRRRRM{The} hierarchical fidelity consideration is shown to be able to  help preserve  moderate-frequency details and thus improve  quality of the produced HR images. 
	
	Besides data fidelity, another important aspect for  MAP based image SR methods is priors on HR images, which are effective in relieving \RRRRRM{ill-posedness} of the problem. \RM{Commonly used  priors \RRRRRM{describing} natural image properties include {sparseness}~\cite{sparsity_2004_CPAM,SparseLinear_2010_IEEE}, \RRRM{spatial smoothness}~\cite{TV_2008_JSC,TV_2005_TIP} and {nonlocal similarity}~\cite{nonlocal1_2008_ECCV}}}, \RM{which help \RRRRRM{produce} more visually pleasant \RRRRRM{HR images}.}
Among those priors, the edge  prior~\cite{EPISR,dai_soft_2007,edge_prior} is a very important one. 
 In contrast to textures that are usually difficult to recover after image degradation,  edges are much easier to detect \RRRRRM{in} LR images and thus  more informative for recovering details of HR images. 
Thus, separating edges from the image signal and modeling them separately would  benefit  image SR  substantially.

Recently, several deep learning based SR methods have been developed, in order to utilize the strong capacity of deep neural networks in modeling complex image contents and  details. The image super-resolution CNN (SRCNN)~\cite{2014_Dong_SRCNN} is the seminal work that has introduced
a deep convolutional network model  to image SR. 
The proposed SRCNN consists of three convolutional layers and is equivalent to performing a sparse reconstruction to generate HR images.
Benefiting from being end-to-end trainable, SRCNN improves the quality of image SR significantly. However, SRCNN only aims at minimizing the MSE loss without exploiting natural image priors and \RRRRRM{suffers} from losing sharp details. 
Following SRCNN, several recent works~\cite{Osendorfer2014,Wang_2015_ICCV} propose to \RRRRRM{embed} sparsity priors into the deep networks for image SR, offering \RRRM{more visually pleasant results}. 
\RRRM{However, much  domain knowledge and extra effort are needed for designing a suitable architecture to model the sparsity priors. A simple and adaptive method to embed various priors into standard CNN networks for image SR is still absent.
} 

\begin{figure}[t] 
	\centering
	\includegraphics[width=8cm]{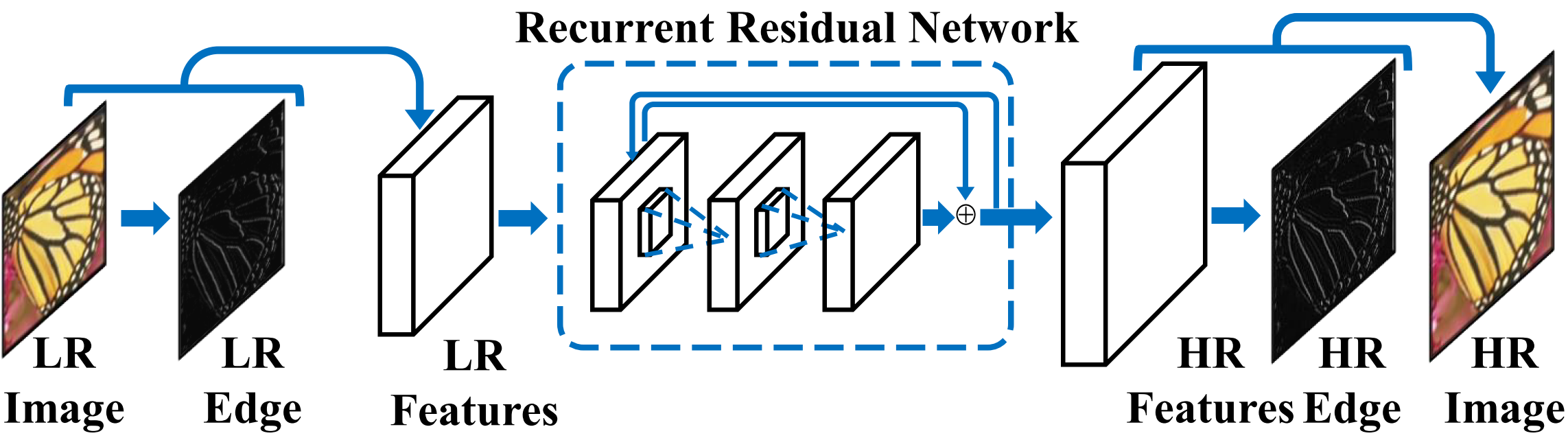} 
	\vspace{-3mm}
	
	\caption{\RRRM{The framework of the proposed \RRRRM{DEGREE} network. The recurrent residual network recovers sub-bands of the HR image features iteratively and edge features are utilized as the guidance in image SR for \RRRRRM{preserving sharp details}.}}
	\vspace{-7mm}
	
	\label{fig:outline} 
\end{figure}

\RM{Motivated by the fact that edge features  can provide valuable guidance for image SR \RsixM{and the success of deep neutral network models}, 
\RM{we propose a Deep Edge Guided REcurrent rEsidual~(DEGREE) network to progressively perform image SR with properly modeled edge priors. 
Instead of trying to predict HR images from LR ones directly, the DEGREE model takes an alternative route and focuses on predicting the \emph{residue} between a pair of LR and HR images, as well as the edges in  HR images. Combining these predictions together give a \RsixM{recovered} HR image \RRRRRM{with} high quality and sharp high-frequency details. \RM{An overview on  the architecture of the DEGREE model is provided in Figure~\ref{fig:outline}}.}
Given an LR image, DEGREE extracts its edge features and takes the features to  predict edges of the HR image via a deep recurrent network.  To recover  details of an  HR image progressively, DEGREE adopts a recurrent residual learning architecture  that  recovers  details \RRRRRM{of} different frequency sub-bands at multiple recurrence stages. Bypass connections are introduced to fuse  recovered results from previous stages and propagate the fusion results to later stages. In addition, adding bypass connections enables a deeper network trained with faster convergence rate.}  


In summary, our contributions to image SR can be summarized as:
\vspace{-2mm}

\begin{enumerate}
	\item We introduce a novel \RRRRM{DEGREE network model} to solve image SR problems.  The DEGREE network  models edge priors and performs image SR recurrently, and improves the quality of produced HR images in a progressive manner. DEGREE is end-to-end trainable and \RRRRRM{effective in exploiting} edge priors for both LR and HR images. To the best of our knowledge, DEGREE is the first recurrent network model with residual learning for recovering HR images.
	\item \RRRRRM{We provide a general framework for embedding natural image priors into image SR, which jointly predicts the task-specific targets and feature maps reflecting specific priors. It is also applicable to other image processing tasks.}
	\item We demonstrate that the recurrent residual learning with bypass structures, designed under the guidance of \RRRM{the sub-band signal reconstruction}, is more effective in image SR than the standard feed forward architecture used in the modern CNN model. DEGREE outperforms well-established baselines significantly on three benchmark datasets and \RRRRRM{provides} new state-of-the-arts.
\end{enumerate}
\vspace{-8mm}

\section{Related Work}

\vspace{-3mm}

Many recent works have exploited  deep learning for solving low level image processing  problems including
image denoising~\cite{2010_Pascal_SDA}, image completion~\cite{2012_Xie_NIPS} and image super-resolution~\cite{2015_Dong_SRCNN}.
Particularly, Dong \textit{et al.}~\cite{SRCNN} proposed a three layer CNN model for image SR through equally performing sparse coding.
Instead of using a generic CNN model, Wang \textit{et al.}~\cite{Wang_2015_ICCV} incorporated the sparse prior into CNN by exploiting a learned iterative shrinkage and thresholding algorithm (LISTA), which provided better reconstruction~performance. 

To address the \RRM{high-frequency} information loss issue in purely minimizing the MSE, \RRM{sub-band decomposition} based  methods propose to \RsixM{recover information at} different frequency \RsixM{bands} of the image signal separately~\cite{sub-band1,singh2014super,Song-ICASSP-2016,sub0,Chatterjee2007}. In~\cite{sub0}, interpolation to \RRM{high-frequency} sub-band images by discrete  wavelet transform (DWT) was performed for image SR. 
In~\cite{Song-ICASSP-2016}, Song \textit{et al.} proposed a joint \RRM{sub-band}-based neighbor-embedding SR with a constraint on each sub-band, achieving more promising SR results. 
%

 Some works also explore how to  preserve edges in \RsixM{application of} image SR, denoising and deblurring.  Total variation (TV)~\cite{tv1,tv2}, focusing on modeling \RsixM{the intensity change} of image signals, was proposed to guide the SR recovery by suppressing the excessive and possibly spurious details in the HR estimation. \RRRM{Bilateral TV (BTV)~\cite{btv1,btv2} was then developed  to preserve sharp edges.} Sparsity prior~\cite{ScSR_2010_TIP,dong_image_2011} constraining the transformation coefficients was introduced to enhance salient features. As a kind of sparsity prior, the gradient prior~\cite{gradient1,gradient2,gradient3} was proposed to enforce the gradient distribution of the denoised image to fit  distribution estimated from the original image. By embedding these regularizations, sharper and finer edges of HR images are restored.
\vspace{-5mm}

\begin{figure}[t] 
	\centering\includegraphics[width=10cm]{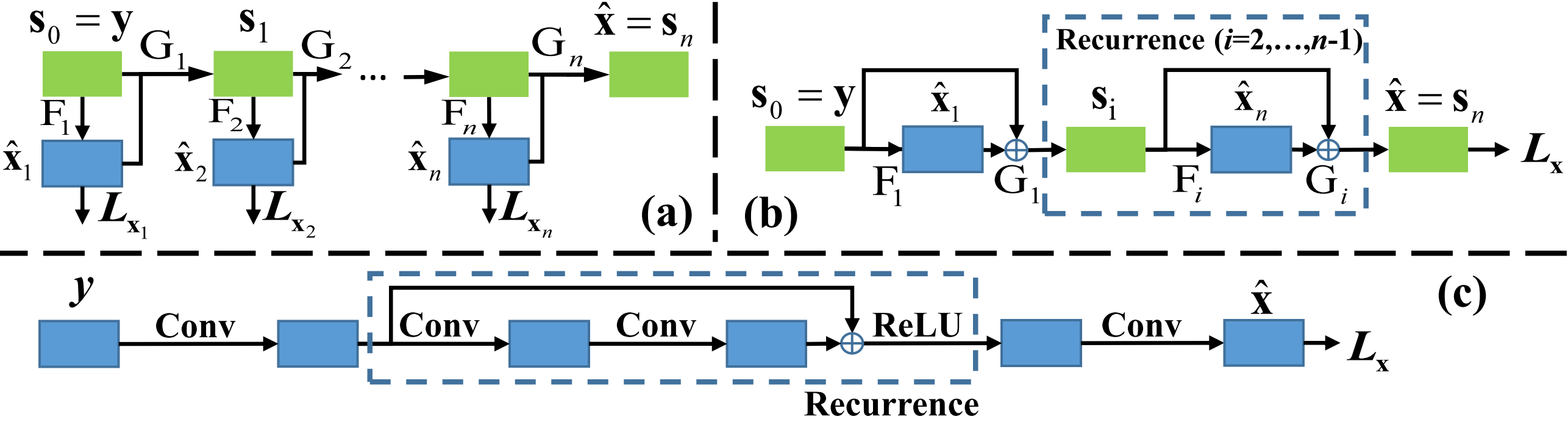} 
	\vspace{-3mm}
	
	\caption{(a) The flowchart of the sub-band \RRM{reconstruction for image super-resolution}. (b) A relaxed version of (a). ${G}_i$ is set as the element-wise summation function.  In this framework, only the MSE loss is used to constrain the recovery. (c) The deep network designed with the intuition of (b). ${G}_i$ is the element-wise summation function and ${F}_i$ is modeled by two layer convolutions.}
	\vspace{-5mm}
	
	\label{fig:subBandDecomp} 
\end{figure}

\section{\RRRRRM{{Deep Recurrent Residual Learning for Image SR}}}
\label{sec:deepResidual}
\vspace{-3mm}

In this section we first review the \RRM{sub-band reconstruction methods}\RRRM{~\cite{sub0,sub-band1}} for image SR. Then we illustrate how to build a recurrent residual network that \RRRRRM{can} learn to perform \RRM{sub-band reconstruction} and recover HR images \RRRRRM{progressively}.
\vspace{-5mm}

\subsection{Sub-Band \RRM{Reconstruction} for Image SR}
\label{sec:subBandDecomp}
\vspace{-2mm}

\RM{In most cases,  quality degradation  of an HR \RsixM{image \textbf{x}} to \RsixM{an LR image \textbf{y}} is \RRRRRM{caused by} blurring and down sampling\RsixM{, and the degradation process can be} modeled as
\vspace{-7mm}

\begin{footnotesize}
\begin{equation}
	\label{equ:degrade}
	\mathbf{y} = DH\mathbf{x}+v,
\end{equation}\end{footnotesize}
\vspace{-5mm}

\noindent where $H$ and $D$ depict the blurring and down-sampling effects respectively. The additive noise in the imaging process is denoted as $v$. Given an observed LR image $ \mathbf{y}$, image SR aims at estimating the original HR $\mathbf{x}$. Most of image SR methods \RsixM{obtain an estimation of HR} by solving the following MAP problem:
\vspace{-1mm}
\begin{footnotesize}
\begin{equation}
	\footnotesize
	\widehat{\mathbf{x}} = \arg\min_{\mathbf{x}}\|DH\mathbf{x}-\mathbf{y}\|_2^2+p(\mathbf{x}),
\end{equation}
\end{footnotesize}
\vspace{-5mm}

\noindent where $p(\cdot)$ is \RRRRRM{a} \RRM{regularization term} induced by \RRRRRM{priors} on $ \mathbf{x} $. However, \RRRRRM{directly} learning \RRRRRM{a one-step} mapping function \RsixM{from \textbf{y} to \textbf{x} usually} ignores some intrinsic properties hidden in different frequency bands of \RRRRRM{$\mathbf{x}$}, such as the high-frequency edge and textural details. This is because the recovery function needs to fit the inverse \RsixM{mapping from the low-frequency component of the LR image to that of the HR one. It} by nature neglects some high-frequency \RsixM{details with small energy}.

\RRRRRM{To address} this problem, a sub-band based \RRM{image reconstruction} method \RRRRRM{is proposed to recover images at different frequency bands separately. It separates the image signal into multiple components of different intrinsic frequencies, which are called sub-bands, and models them \RsixM{individually}. In this way,}}
the sub-bands with small energy \RsixM{can still} gain \RRRRRM{sufficient} ``attention'' and sharper image details \RRRRRM{can be} preserved \RsixM{during image SR}.
Formally, let $\mathbf{y}_i$ be the $i$-th sub-band of the LR image $\mathbf{y}$ out of in total $n$ sub-bands, \RsixM{\textit{i.e.}, $\mathbf{y} = \sum_{i=1}^{n}\mathbf{y}_i$. $\mathbf{y}_i$} is used for estimating the $i$-th \RsixM{corresponding} sub-band  $\mathbf{x}_i$ of the HR image $\mathbf{x}$. 
The \RRM{sub-band} based method recovers different sub-bands \RsixM{individually and outputs the recovered HR image} as follows,
\vspace{-4mm}

\begin{footnotesize}
\begin{equation}
	\label{equ:MAP}
	\widehat{\mathbf{x}}_i =\arg\min_{\mathbf{x}_i}\|DH\mathbf{x}_i-\mathbf{y}_i\|_2^2+p(\mathbf{x}_i), i =1,2,...,n; \text{ } \widehat{\mathbf{x}} = \sum_{i=1}^{n}\widehat{\mathbf{x}}_i.
\end{equation}
\end{footnotesize}
\vspace{-3mm}

\noindent \RsixM{However, recovering each sub-band separately in (\ref{equ:MAP}) neglects the dependency across sub-bands. To fully model the dependencies both in the corresponding sub-bands and across sub-bands, we relax (\ref{equ:MAP}) into a progressive recovery process. It performs an iterative sub-band recovery implicitly and utilizes the useful information from lower-frequency sub-bands to recover higher-frequency~ones.}

\RsixM{For ease of explanation,} \RRRRRM{we introduce} an auxiliary signal $\mathbf{s}_i$ that approximates the signal $\mathbf{x}$ up to the $i$-th sub-band, \emph{i.e.}, $\mathbf{s}_i = \sum_{j=1}^{i}\widehat{\mathbf{x}}_j$. Then, the sub-band \RRRRRM{image} $\mathbf{x}_i$ and \RRRRRM{HR image} $\mathbf{x}$ can be \RRRRRM{estimated} through recovering $\widehat{\mathbf{x}}_i$ and $\mathbf{s}_i$ \RsixM{progressively}. \RsixM{We} \RRRRRM{here} use ${F}_i(\cdot)$ and ${G}_i(\cdot)$ to denote the \RsixM{generating functions of} $\mathbf{s}_i$ and $\widehat{\mathbf{x}}_{i}$ \RsixM{respectively, \emph{i.e.}},
\vspace{-6mm}

\begin{footnotesize}
\begin{align}
	\label{lab:fix-sub-band}
	\widehat{\mathbf{x}}_{i} = {F}_i(\mathbf{s}_{i-1}),\text{ }
	\mathbf{s}_{i} ={G}_i(\widehat{\mathbf{x}}_i,\mathbf{s}_{i-1}),
\end{align}
\end{footnotesize}
\vspace{-6mm}

\noindent where $\mathbf{s}_0 = \mathbf{y}$ \RsixM{is the input LR image} and $\mathbf{s}_n$ eventually re-produces \RsixM{the HR image} $\mathbf{x}$.  Figure \ref{fig:subBandDecomp}(a) gives an overall illustration on this process. \RsixM{The functions} ${F}_i$ and ${G}_i$ \RsixM{usually take} linear transformations as advocated in~\cite{sub-band1,singh2014super,Song-ICASSP-2016}. \RRRM{${F}_i$ \RsixM{learns to recover} high frequency detail, estimating the $i$-th sub-band \RsixM{component} based on the accumulated recovered results \RsixM{from} previous $(i-1)$ sub-bands. ${G}_i$ \RsixM{fuses} $\widehat{\mathbf{x}}_i$ and $\mathbf{s}_{i-1}$ \RsixM{in order to balance} different sub-bands.} \RsixM{In the figure,} \RRRRM{$\pmb{L}_{\mathbf{x}_i}$ is the loss term corresponding to the data fidelity \RsixM{in} (\ref{equ:MAP}).} The progressive sub-band recovery can be learned in a supervised way~\cite{sub-band1,sub-band2}, where the ground truth sub-band signal $\mathbf{x}_i$ is generated by applying \RsixM{band filters}
on $\mathbf{x}$. In our proposed method, we choose the element-wise summation function to model ${G}_i$ in the proposed network, following the additive assumption for the sub-bands of the image signal that is generally implied in previous methods~\cite{sub0,sub-band1}. 
\vspace{-4mm}

\subsection{Learning Sub-Band Decomposition by Recurrent Residual Net}
\label{sec:subBandNetwork}
\vspace{-2mm}

The sub-band paradigm mentioned above \RsixM{learns to recover HR images} \RsixM{through minimizing} a hierarchical loss generated by applying \RsixM{hand-crafted} frequency domain filters, as shown in \RsixM{Figure \ref{fig:subBandDecomp}(a). \RsixM{However, this paradigm} suffers from \RsixM{following} two limitations. First, it does not provide an end-to-end trainable framework. Second, it suffers from the heavy dependence on the choice of the frequency filters. A bad choice of the filters would severely limit its capacity of modeling the correlation between different sub-bands, and recovering the HR~$\mathbf{x}$.}

To handle \RsixM{these two problems, by employing a summation function as $G_i$, we reformulate the recover process in (\ref{lab:fix-sub-band}) into:}
\vspace{-7mm}

\RsixM{
\begin{footnotesize}
	\begin{align}
		\label{lab:whole-sub-band}
		\mathbf{s}_{i} = \mathbf{s}_{i-1} + {F}_i(\mathbf{s}_{i-1}).
	\end{align}
\end{footnotesize}
}
\vspace{-7mm}

\noindent\RsixM{In this way, the intermediate estimation $\widehat{\mathbf{x}}_{i}$ is not necessary to estimate explicitly. An end-to-end training paradigm can then be constructed as shown in Figure \ref{fig:subBandDecomp}(b). The MSE loss $\pmb{L}_{\mathbf{x}}$ imposed at the top layer is the only constraint on ${\widehat{\mathbf{x}}}$ for the HR prediction. Motivated by (\ref{lab:whole-sub-band}) and Figure~\ref{fig:subBandDecomp}(b), we further  propose a recurrent residual learning network whose architecture is shown in Figure~\ref{fig:subBandDecomp}(c). To increase the modeling ability, $F_i$ is parameterized by two layers of convolutions. To introduce nonlinearities into the network, $G_i$ is modeled by an element-wise summation connected with a non-linear rectification.
Training the network to minimize the MSE loss gives the functions $F_i$ and $G_i$ adaptive to the training data.
Then, we stack $n$ recurrent units into a deep network to perform a progressive sub-band recovery. Our proposed recurrent residual network follows the intuition of gradual sub-band recovery process. 
The proposed model is equivalent to balancing the contributions of each sub-band recovery. Benefiting from the end-to-end training, such deep sub-band learning is more effective than the traditional supervised sub-band recovery. Furthermore, the proposed network indeed has the ability to recover the sub-bands of the image signal recurrently, as validated in Section~\ref{exp:Visualization}}.
\vspace{-4mm}

\section{\M{DEGREE Network for Edge Preserving SR}}
\vspace{-3mm}

We have presented how to construct a recurrent residual network to perform deep \RRM{sub-band} learning. In this section, we proceed to explain how to embed the edge prior into the recurrent residual network, in order to predict high-frequency details \RsixM{better for} image SR. 
\vspace{-3mm}

\subsection{Edge \RsixM{Extraction}}
\vspace{-1mm}

\RsixM{An} HR image $\mathbf{x}$ \RsixM{can be} separated into low-frequency and high-frequency components, as $\mathbf{x} =\mathbf{x}_L+\mathbf{x}_H$,
%
\RsixM{where the} high-frequency component $\mathbf{x}_H$ contains subtle details \RRRRRM{of the image}, such as edges and textures. Patterns \RRRRRM{contained} in $\mathbf{x}_H$ are \RRRRRM{usually} irregular and have smaller magnitude compared with $\mathbf{x}_L$. Thus, in image degradation, the component of $\mathbf{x}_H$ is more fragile and \RsixM{easier} to be \RRM{corrupted}, \RsixM{which is also difficult to recover.} \RsixM{To better recover $\mathbf{x}_H$}, we propose to extract extra prior knowledge about $\mathbf{x}_H$ from \RsixM{the} LR image $\mathbf{y}$ \RsixM{as a build-in component in the deep recurrent residual network to regularize the recovery process}. Among all the statistical priors about natural images, edge is one of the most \RRRRRM{informative} priors.
Therefore, we \RsixM{propose to model edge priors and develop a deep edge guided recurrent residual network, which is introduced in the following section. However, our proposed network architecture can also embed other statistical priors extractable from LR inputs for image SR.}
\RsixM{To extract edges, we} first apply an off-the-shelf  edge detector (such as the Sobel one) on $\mathbf{y}$ and \RRRM{$\mathbf{x}$} to get its high-frequency component $\mathbf{y}_H$ and \RRRM{$\mathbf{x}_H$}. Then we \RsixM{train} the model \RsixM{to} predict $\mathbf{x}_H$ based on both $\mathbf{y}$ and $\mathbf{y}_H$. \RsixM{Please note that $\mathbf{x}_H$ is the high-frequency residual of~$\mathbf{x}$.}

\begin{figure}[t]
	\centering\includegraphics[width=10cm]{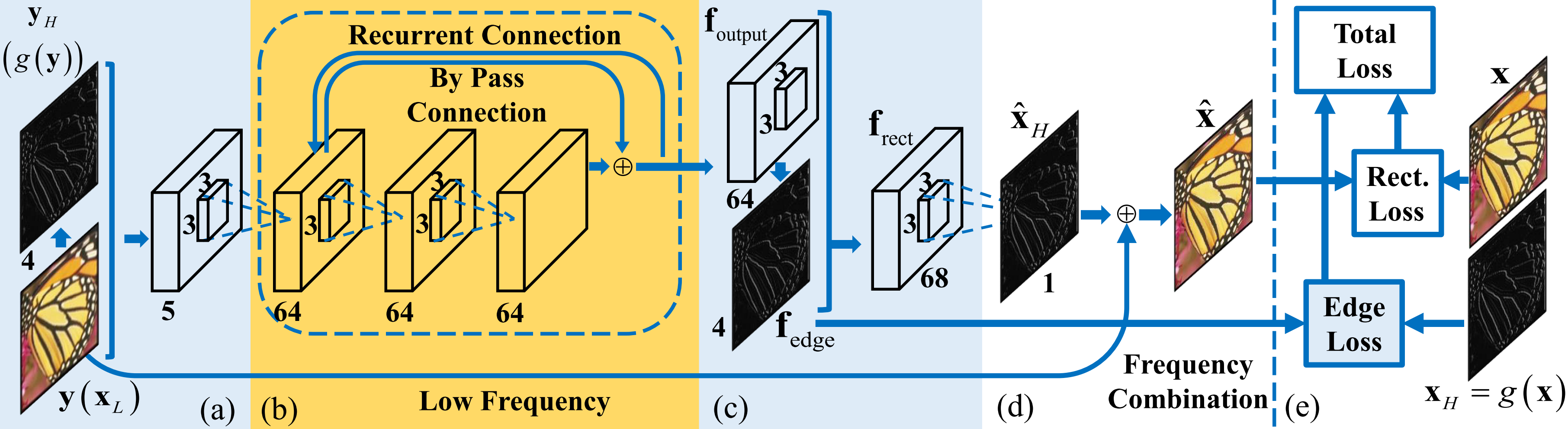} 
	\vspace{-2mm}
	
	\caption{ The architecture of the \RRRRM{DEGREE network for image SR.}
	\RRM{(a) The LR edge maps $\mathbf{y}_H(g(\mathbf{y}))$ of the LR image are part of the input features. (b) Recurrent residual learning network for sub-band recovery. (c) Part of the feature maps $\mathbf{f}_{\text{edge}}$ in the penultimate layer aim at generating HR edges. (d) Combining the \RRM{high-frequency} estimation and the LR image by $\widehat{\mathbf{x}}=\mathbf{x}_L+\widehat{\mathbf{x}}_H$. (e) The total loss is the combination of the edge loss and reconstruction loss, which constrain the recovery of HR edges and HR images respectively. Our main contributions, the edge guidance and recurrent residual learning, are highlighted with blue and \RRRRM{orange} colors.}
	}
	\label{fig:framework_fixed_edge} 
	\vspace{-5mm}
	
\end{figure}
\vspace{-4mm}

\subsection{DEGREE Network}
\vspace{-1mm}


We propose an end-to-end \RRRRM{trainable deep edge guided recurrent residual network (DEGREE) for image SR.} The network is constructed based on the following two intuitions. First, as we have demonstrated, a recurrent residual network is capable of learning \RRM{sub-band decomposition and reconstruction} for image SR. Second, \RsixM{modeling} \RsixM{edges extracted} from the LR image \RsixM{would} benefit recovery of details in the HR image. An overview on the architecture \RsixM{of the proposed DEGREE network is given in Figure~\ref{fig:framework_fixed_edge}. As shown in the figure,} \RsixM{DEGREE contains following components.} \textbf{a) LR Edge Extraction}.
An edge map of the input LR image is extracted by \RsixM{applying} a hand-crafted edge detector and is fed into the network together with the raw LR image, as shown in Figure \ref{fig:framework_fixed_edge}(a). \textbf{b) Recurrent Residual Network.}
The mapping function from LR images to HR images is modeled by the recurrent residual network as introduced in Section~\ref{sec:subBandNetwork}, 
\RsixM{Instead of predicting the HR image directly, DEGREE recovers the residual image at different frequency sub-bands progressively and combine them into the HR image, as shown in Figure~\ref{fig:framework_fixed_edge}(b).}
\textbf{c) HR Edge Prediction}. \RsixM{DEGREE produces} convolutional feature maps in the penultimate layer, \RsixM{part of which ($\mathbf{f}_{\text{edge}}$)} are used to reconstruct the edge maps \RsixM{of the} HR images and provide extra knowledge for reconstructing the HR images, as shown  in Figure~\ref{fig:framework_fixed_edge}(c). 
\textbf{d) \RsixM{Sub-Bands Combination For Residue}.} \RRRRM{Since the  LR image contains necessary \RRM{low-frequency} details,  \RsixM{DEGREE only focuses} on recovering the high-frequency component, especially several high-frequency sub-bands of the HR image, which are the differences or \emph{residue} between the HR image and the input LR image. Combining the estimated residue with sub-band signals and the LR image gives an HR image, as shown in Figure \ref{fig:framework_fixed_edge}(d).} \textbf{e) \RsixM{Training} Loss.} We consider the reconstruction loss of both the HR image and HR edges simultaneously \RsixM{for} training \RsixM{DEGREE} as shown in Figure~\ref{fig:framework_fixed_edge}(e). 
We now explain each individual part of the proposed network in details. 
\vspace{-6mm}

\subsubsection{\M{Recurrent \RsixM{Residual} Network}}
The recurrent residual network aims to refine SR images progressively \RsixM{through producing the residue image at different frequency}. We follow the notations in Section~\ref{sec:deepResidual}. \RsixM{To provide a formal description}, let $\mathbf{f}_{\text{in}}^k$ denote the input feature map for the recurrent sub-network at the $k$-th time step. 
The output feature map $\mathbf{f}_{\text{out}}^k$ of \RRRRRM{the} recurrent sub-network is progressively updated as follows,
\vspace{-5mm}

\RRRRRM{
\begin{footnotesize}
\begin{align}
\mathbf{f}_{\text{out}}^k = \max{\left(0, \mathbf{W}_{\text{mid}}^k*\mathbf{f}_{\text{mid}}^k+\mathbf{b}_{\text{mid}}^k\right)}+\mathbf{f}_{\text{in}}^k, \text{ with }\mathbf{f}_{\text{mid}}^k = \max{\left(0, \mathbf{W}_{\text{in}}^k*\mathbf{f}_{\text{in}}^k+\mathbf{b}_{in}^k\right)},
\end{align}
\end{footnotesize}}
\vspace{-5mm}

\noindent where $\mathbf{f}_{\text{in}}^k = \mathbf{f}_{\text{out}}^{k-1}$ is the output features by the  recurrent sub-network at $(k-1)$-th time step.
Please note the by-pass connection here between $\mathbf{f}_{\text{in}}^k$  and $\mathbf{f}_{\text{out}}^k$. 
In the context of sub-band reconstruction, the feature map $\mathbf{f}_{\text{out}}^k$ can be viewed as the recovered $k$-th sub-band of the image signal. Let $K$ be the total recurrence number of the sub-networks. Then, the relation between $\mathbf{f}_{\text{in}}^1$, $\mathbf{f}_{\text{out}}^K$ and  the overall network is
\vspace{-7mm}

\begin{footnotesize}
\begin{align}
&\mathbf{f}_{\text{in}}^1 = \max(0, \mathbf{W}_{\text{input}}*\mathbf{f}_{\text{input}}+\mathbf{b}_{\text{input}}), \\
&\mathbf{f}_{\text{output}} = \mathbf{f}_{\text{out}}^K, \nonumber
\end{align}
\end{footnotesize}
\vspace{-7mm}

\noindent \RRM{where $\mathbf{W}_{\text{input}}$ and $\mathbf{b}_{\text{input}}$ denote the filter parameter and basis of the convolution layer  before the recurrent sub-network.} \RRRRM{Thus, $\mathbf{f}_{\text{output}}$ is the output features of the recurrent residual network, which are used to reconstruct both the HR features and images. }
	
\vspace{-5mm}

\subsubsection{\M{Edge Modeling}}
\label{sec:edgePrediction}
\RsixM{We here illustrate how to embed the edge information into the proposed deep network.}
\RsixM{This can also generalize to modeling other natural image priors.}
In particular, the proposed network takes edge features extracted from the LR image as \RsixM{another} input, and aims to predict edge maps of the HR image as a part of \RsixM{its} output features \RsixM{which are then} used for recovering the HR~image. 

The input feature $\mathbf{f}_{\text{input}}$ to the network is a concatenation of  the  raw LR image $\mathbf{y}$ and its edge map $ g(\mathbf{y})$,
\vspace{-3mm}

\begin{footnotesize}
\begin{equation}
\mathbf{f}_{\text{input}} = \left[\mathbf{y}, g(\mathbf{y})\right].
\end{equation}
\end{footnotesize}
\vspace{-5mm}

\noindent To recover the HR image, \RsixM{DEGREE} outputs two types of features at its  penultimate layer.  One is for HR image recovery and the other one is for edge prediction in the HR image. 
More specifically, let $\mathbf{f}_{\text{output}}$ denote  the features used to reconstruct  HR images and let $\mathbf{f}_{\text{edge}}$ denote the edge feature computed~by
\vspace{-2mm}

\begin{footnotesize}
\begin{equation}
\mathbf{f}_{\text{edge}} = \max{\left(0, \mathbf{W}_{\text{edge}}*\mathbf{f}_{\text{output}}+\mathbf{b}_{\text{edge}}\right)},
\end{equation}
\end{footnotesize}
\vspace{-4mm}

\noindent where $\mathbf{W}_{\text{edge}}$ and $\mathbf{b}_{\text{edge}}$ are the filter and the bias of the convolution layer to  predict the HR edge map. Thus, the features $\mathbf{f}_{\text{rect}}$ in the penultimate layer for reconstructing the HR image with the edge guidance are given as follows,
\vspace{-3mm}

\begin{footnotesize}
\begin{equation}
\mathbf{f}_{\text{rect}} = \left[ \mathbf{f}_{\text{output}}, \mathbf{f}_{\text{edge}}\right].
\end{equation}
\end{footnotesize}
\vspace{-10mm}

\subsubsection{\M{Sub-Bands Combination}}
In  sub-band based image SR methods, the \RRM{low-frequency} and high-frequency components of an image signal are usually extracted at different parts in a hierarchical decomposition of the signal.
\RsixM{DEGREE} network also models the \RRM{low-frequency} and \RRM{high-frequency} components of an image \RsixM{jointly}. 
Denote the \RRM{high-frequency} and \RRM{low-frequency} components of an HR image $\mathbf{x}$  as $\mathbf{x}_H$ and $ \mathbf{x}_L$  respectively. We have $\mathbf{x} = \mathbf{x}_H + \mathbf{x}_L$.
Here, we use the notation  $\mathbf{y}$ to denote both the original LR image and its up-scaled version of the same size as $\mathbf{x}$,  if it causes no confusion. 
Obviously, $\mathbf{y}$ is  a good estimation of the low frequency component $\mathbf{x}_L$ of the HR image $\mathbf{x}$.
The retained high-frequency component $\mathbf{y}_H$ of $\mathbf{y}$, \emph{i.e.}, the edge map of $\mathbf{y}$, is  estimated by applying an edge extractor (we use Sobel) onto  $\mathbf{y}$.
In our proposed \RsixM{DEGREE} network, as shown in Figure~\ref{fig:framework_fixed_edge}, the low-frequency component $\mathbf{x}_L \approx \mathbf{y}$ is directly passed to the last layer and combined with the predicted high-frequency image $\widehat{\mathbf{x}}_H$  to produce an estimation $\mathbf{\widehat{x}}$ of the  HR image $\mathbf{x}$: $\mathbf{\widehat{x}} = \mathbf{x}_L + \mathbf{\widehat{x}}_H$.
Here, $\mathbf{\widehat{x}}_H$, an estimation of the high-frequency component $\mathbf{x}_H$, is generated by
\vspace{-3mm}

\begin{footnotesize}
\begin{equation}
\mathbf{\widehat{x}}_H = \max{\left(0, \mathbf{W}_{\text{rect}}*\mathbf{f}_{\text{rect}}+\mathbf{b}_{\text{rect}}\right)},
\end{equation}
\end{footnotesize}
\vspace{-5mm}

\noindent where 
$\mathbf{f}_{\text{rect}}$ is the features learned in the penultimate layer to reconstruct $\mathbf{x}_H$. The filters and biases involved in the layer are denoted as $\mathbf{W}_{\text{rect}}$ and $\mathbf{b}_{\text{rect}}$.
\vspace{-5mm}

\subsubsection{\M{Training}}
Let $\mathbf{F}(\cdot)$ represent the learned network for recovering the HR image $\mathbf{x}$ based on the input LR image $\mathbf{y}$ and the LR edge map $\mathbf{y}_H$. Let $\mathbf{F}_{\text{edge}}(\cdot)$ denote the learned HR edge predictor which outputs $\mathbf{f}_{\text{edge}}$. We use $\mathbf{\Theta}$ to collectively denote all the parameters of the network,
\vspace{-6mm}

\begin{footnotesize}
\begin{equation}
\mathbf{\Theta}=\left\{\mathbf{W}_{\text{input}},\mathbf{b}_{\text{input}},\mathbf{W}_{\text{in}},\mathbf{b}_{\text{in}},\mathbf{W}_{\text{mid}},\mathbf{b}_{\text{mid}},\mathbf{W}_{\text{edge}},\mathbf{b}_{\text{edge}},\mathbf{W}_{\text{rect}},\mathbf{b}_{\text{rect}}\right\}.
\end{equation}
\end{footnotesize}
\vspace{-6mm}

\noindent Given $n$ pairs of  HR and LR images $\left\{(\mathbf{x}_i, \mathbf{y}_i )\right\}_{i=1}^n$ for training, 
we first extract the  \RRM{high-frequency} components of LR and HR images,  $\left\{\mathbf{y}_{i,H}\right\}$ and $\left\{\mathbf{x}_{i,H}\right\}$, by applying Sobel operator on the image $\mathbf{x}_i$ and $\mathbf{y}_i$ respectively. 
We adopt the following joint mean squared error (MSE)  to train the network parameterized by $\mathbf{\Theta}$ such that it can jointly estimate the HR images and HR edge maps:
\vspace{-5mm}

\begin{footnotesize}
\begin{align}
&L(\mathbf{\Theta})=\frac{1}{n}\sum_{i=1}^{n}(||\mathbf{F}\left(\mathbf{y}_{i},\mathbf{y}_{i,H},\mathbf{x}_{i},\mathbf{x}_{i,H};\mathbf{\Theta}\right)-\mathbf{x}_i||^2+\lambda||\mathbf{F}_{\text{edge}}\left(\mathbf{y}_{i},\mathbf{y}_{i,H},\mathbf{x}_{i},\mathbf{x}_{i,H};\mathbf{\Theta}\right)-\mathbf{x}_{i,H}||^2). \nonumber
\end{align}
\end{footnotesize}
\vspace{-4mm}

\noindent Here $\lambda$ is a trade-off parameter that balances  importance of the data fidelity term and the edge prior term. We empirically set $\lambda$ as $1$ throughout the paper because we observe that our method performs similarly for different values of $\lambda$ in a large range, as mentioned in Section~\ref{Exp} and validated in supplementary~material. 
\vspace{-5mm}

\section{Experiments}
\label{Exp}
\vspace{-3mm}

\subsubsection{Datasets}
Following the experimental setting in [43] and [44], we compare the proposed method with recent SR methods on three popular benchmark datasets: Set5~\cite{set5_ref}, Set14~\cite{set14_ref} and BSD100~\cite{bsd100_ref} with scaling factors of 2, 3 and 4. The three datasets contain 5, 14 and 100 images respectively. Among them, the Set5 and Set14 datasets are commonly used  for evaluating  traditional image processing methods, and the BSD100 dataset contains 100 images with diverse natural scenes.
\RRRM{We train our model \RRRRRM{using} a training set created in~\cite{yang_image_2010}, which contains 91 images.} \RRRRRM{For fair comparison with other methods \cite{Wang_2015_ICCV}, we do not train our models with a larger dataset.}
We either do not use any ad-hoc post-processing.
\vspace{-6mm}

\subsubsection{Baseline Methods}
\RRM{We compare our \RRRRM{DEGREE SR network~(DEGREE)} with Bicubic interpolation and the following six state-of-the-art SR methods: ScSR (Sparse coding)~\cite{ScSR_2010_TIP}, A+ (Adjusted Anchored Neighborhood Regression)~\cite{Aplus_2015_ACCV}, SRCNN~\cite{SRCNN}, TSE-SR (Transformed Self-Exemplars)~\cite{Huang-CVPR-2015}, CSCN (Deep Sparse Coding)~\cite{Wang_2015_ICCV} and JSB-NE (Joint Sub-Band Based Neighbor Embedding)~\cite{Song-ICASSP-2016}.
It is worth noting that CSCN and JSB-NE are the most recent deep learning and sub-band recovery based image SR methods respectively. }
\vspace{-5mm}

\subsubsection{Implementation Details}
We evaluate our proposed model with 10 and 20 layers respectively. The bypass connections are set with an interval of 2 convolution layers, as \RRRRRM{illustrated} in Figure \ref{fig:framework_fixed_edge}. The number of channels in each convolution layer is fixed as $64$ and the filter size is set as $3\times3$ with a padding size of $1$. All these settings are  consistent with the  one used in \cite{deepResidual}. The edge extractor is applied along four directions \RRRM{(up-down, down-up, left-right and right-left)} \RRRRRM{for extracting edge maps}. Following the experimental setting in \cite{SRCNN}, we generate LR images by applying Bicubic interpolation on the HR images. The training and \RRRRRM{validation} images are cropped into small sub-images with a size of $33 \times 33$ pixels. \RRRM{We use flipping (up-down and left-right) and clockwise rotations ($0^{\circ}, 90^{\circ}, 180^{\circ}$ and $270^{\circ}$) \RRRRRM{for} data augmentation. \RRRRRM{For} each \RRRRRM{training} image, 16 \RRRRRM{augmented} images are generated.} The \RRRRRM{final} training set \RRRRRM{contains} around 240,000 sub-images. The weighting parameter $\lambda$ for balancing the losses is \RRRRRM{empirically} set as $1$. \RRRRM{We empirically \RRRRRM{show} that the DEGREE network is robust to the choice of $\lambda$ in the supplementary material and the best performance is \RRRRRM{provided} by setting $\lambda \le 1$.} Following the common practice in many previous methods, we only \RRRRRM{perform} super-resolution in the luminance channel (in YCrCb color space).  The other two chrominance channels are bicubically upsampled  for displaying the \RRRRRM{results}. We train our model on the {Caffe} platform \cite{jia2014caffe}. Stochastic gradient descent~(SGD)  with  standard back-propagation is used for training the model. In particular, in the optimization we set momentum  as 0.9, the initial learning rate as $0.0001$ and change it to $0.00001$ after $76$ epochs. We only allow at most $270$ epochs.
\vspace{-4mm}

\subsection{Objective Evaluation}

We use DEGREE-1 and DEGREE-2 to denote two versions of the proposed model when we report the results. DEGREE-1 has \RRRRRM{10} layers and 64 channels, and DEGREE-2 has \RRRRRM{20} layers and 64 channels. The quality of the HR images produced by different SR methods is measured by the Peak Signal-to-Noise Ratio (PSNR)~\cite{psnr_ref} and the perceptual quality metric Structural SIMilarity (SSIM)~\cite{ssim_ref}, which are two widely used metrics in image processing. The results of our proposed DEGREE-1 and DEGREE-2  as well as the baselines are given in Table~\ref{tab:objective}. 
\vspace{-8mm}

\begin{table}[!ht] 
	{\scriptsize
		\centering
		\caption{Comparison among different image SR methods on three test datasets with three  scale factors ($\times 2$, $\times 3$ and $\times 4$). The bold numbers denote the best performance and the underlined numbers denote the second best performance. The performance gain of DEGREE-2 over the  best baseline results is shown in the last row.}
		\label{tab:objective}
		\vspace{-3mm}
		
		\begin{tabular}{l|c|c|c|c|c|c|c|c|c|c}
			\multicolumn{2}{c|}{Dataset}        & \multicolumn{3}{c|}{Set5} & \multicolumn{3}{c|}{Set14} & \multicolumn{3}{c}{BSD100} \\
			\hline
			Method & Metric & $\times$2    & $\times$3    & $\times$4    & $\times$2    & $\times$3    & $\times$4    & $\times$2    & $\times$3    & $\times$4 \\
			\hline
			\multirow{2}[3]{*}{Bicubic} & PSNR  & 33.66  & 30.39  & 28.42  & 30.13  & 27.47  & 25.95  & 29.55  & 27.20  & 25.96 \\
			& SSIM  & 0.9096  & 0.8682  & 0.8105  & 0.8665  & 0.7722  & 0.7011  & 0.8425  & 0.7382  & 0.6672  \\	
			\hline
			\multirow{2}[3]{*}{ScSR} & PSNR  & 35.78  & 31.34  & 29.07  & 31.64  & 28.19  & 26.40  & 30.77  & 27.72  & 26.61  \\
			& SSIM  & 0.9485  & 0.8869  & 0.8263  & 0.8990  & 0.7977  & 0.7218  & 0.8744 & 0.7647  & 0.6983  \\
			\hline	
			\multirow{2}[3]{*}{A+} & PSNR  & 36.56  & 32.60  & 30.30  & 32.14  & 29.07  & 27.28  & 30.78  & 28.18  & 26.77  \\
			& SSIM  & 0.9544  & 0.9088  & 0.8604  & 0.9025  & 0.8171  & 0.7484  & 0.8773  & 0.7808  & 0.7085  \\
			\hline
			\multirow{2}[3]{*}{TSE-SR} & PSNR  & 36.47  & 32.62  & 30.24  & 32.21  & 29.14  & 27.38  & 31.18  & 28.30  & 26.85  \\
			& SSIM  & 0.9535  & 0.9092  & 0.8609  & 0.9033  & 0.8194  & 0.7514  & 0.8855  & 0.7843  & 0.7108  \\
			\hline
			\multirow{2}[3]{*}{JSB-NE} & PSNR  &  36.59 & 32.32  & 30.08  & 32.34  & 28.98  & 27.22  & 31.22  & 28.14  &  26.71 \\
			& SSIM  & 0.9538  & 0.9042  &  0.8508 & 0.9058  & 0.8105  & 0.7393  &  0.8869 & 0.7742  &  0.6978 \\			
			\hline
			\multirow{2}[3]{*}{CNN} & PSNR  & 36.34  & 32.39  & 30.09  & 32.18  & 29.00  & 27.20  & 31.11  & 28.20  & 26.70  \\
			& SSIM  & 0.9521  & 0.9033  & 0.8530  & 0.9039  & 0.8145  & 0.7413  & 0.8835  & 0.7794  & 0.7018  \\
			\hline
			\multirow{2}[3]{*}{CNN-L} & PSNR  & 36.66  & 32.75  & 30.49  & 32.45  & 29.30  & 27.50  & 31.36  & 28.41  & 26.90  \\
			& SSIM  & 0.9542  & 0.9090  & 0.8628  & 0.9067  & 0.8215  & 0.7513  & 0.8879  & 0.7863  & 0.7103  \\
			\hline
			\multirow{2}[3]{*}{CSCN} & PSNR  & 36.88  & 33.10  & 30.86  & 32.50  & 29.42  & 27.64  & 31.40  & 28.50  & 27.03  \\
			& SSIM  & 0.9547  & 0.9144  & 0.8732  & 0.9069  & 0.8238  & 0.7573  & 0.8884  & 0.7885  & 0.7161  \\
			\hline
			\multirow{2}[3]{*}{DEGREE-1} & PSNR  & \underline{37.29} & \underline{33.29}  & \underline{30.88} & \underline{32.87} & \underline{29.53}  & \underline{27.69}  & \underline{31.66} & \underline{28.59} & \underline{27.06} \\
			& SSIM  & \underline{0.9574} & \underline{0.9164}  & \underline{0.8726} & \underline{0.9103} & \underline{0.8265} & \underline{0.7574}  & \textbf{0.8962} & \underline{0.7916}  & \textbf{0.7177} \\
			\hline
			\multirow{2}[3]{*}{DEGREE-2} & PSNR  & \textbf{37.40} & \textbf{33.39 } & \textbf{31.03 } & \textbf{32.96 } & \textbf{29.61 } & \textbf{27.73 } & \textbf{31.73 } & \textbf{28.63 } & \textbf{27.07 } \\
			& SSIM  & \textbf{0.9580 } & \textbf{0.9182 } & \textbf{0.8761 } & \textbf{0.9115 } & \textbf{0.8275 } & \textbf{0.7597 } & \underline{0.8937} & \textbf{0.7921 } & \textbf{0.7177 } \\
			\hline
			\multirow{2}[3]{*}{Gain} & PSNR  & 0.52  & 0.29  & 0.17  & 0.46  & 0.19  & 0.09  & 0.26  & 0.13  & 0.04  \\
			& SSIM  & 0.0033  & 0.0038  & 0.0029  & 0.0046  & 0.0037  & 0.0025  & 0.0053  & 0.0036  & 0.0016  \\
		\end{tabular}
		\vspace{-8mm}
		
		\label{tab:obj}
	}
\end{table}	

From the table, it can be seen  that the our proposed DEGREE models consistently outperform those well-established baselines with significant performance gains. DEGREE-2 performs the best for all the three scaling factors on the three datasets, except for the setting of $\times 2$ on  BSD100 in terms of SSIM, where  DEGREE-1 performs the best. Comparing the performance of DEGREE-1 and DEGREE-2 clearly demonstrates that increasing the depth of the network indeed improves the performance, \RRRRRM{but we observe} that \RRRRRM{further} increasing the depth leads to no performance gain. We also list the concrete performance gain brought by the proposed DEGREE model over the state-of-the-art (CSCN). One can observe that when enlarging the image by a factor of 2, our proposed method can further improve the state-of-the-art performance with a margin up to $0.52$ (PSNR) and $0.0033$ (SSIM) on Set5. For other scaling factors, our method also consistently provides \RRRRRM{better} performance. For example, on \RRRRRM{the} Set5 dataset, DEGREE-2 improves the performance by $0.29$ and $0.17$  for $\times 3$  and $\times 4$ settings respectively. 
Our \RRRRRM{models} are more competitive for a small scale factor. This might be because edge features are more salient and are easier to be predicted in small scaling enlargements. This is also consistent with the observation made for the gradient statistics  in the previous edge-guided SR method~\cite{edge_prior}.
\vspace{-4mm}

\subsection{Subjective Evaluation}
\vspace{-1mm}

We also present some visual results in Figures~\ref{fig:butterfly},~\ref{fig:86000} and \ref{fig:223061} to investigate how the methods perform in terms of visual quality. \RRRM{These results are generated by \RRRRRM{our} proposed network with 20 layers, \RRRRRM{\emph{i.e.} DEGREE-2.}} Since our method is significantly better than baselines for the scaling factor of 2, here we in particular focus on comparing the visual quality of produced images with  larger scaling factors.  Figure~\ref{fig:butterfly} displays the SR results on the image of \textit{Butterfly} from  Set5 for $\times 4$ enlargement. From the figure, one can observe that the results generated by A+, SRCNN and JSB-NE contain artifacts or blurred details. CSCN provides fewer artifacts. But there are \RRRRRM{still} a few remained, \RRRM{such as the corners of yellow and white plaques as shown in the enlarged local result in Figure~\ref{fig:butterfly}}. Our method generates a more visually pleasant image with clean details and sharp edges. For the  image \emph{86000} from BSD100, as shown in Figure~\ref{fig:86000}, our method produces an image with the cleanest window boundary. For the image \emph{223061} from BSD100 in Figure~\ref{fig:223061}  that contains a lot of edges and texture, most of methods generate the results with severe artifacts. Benefiting from explicitly exploiting the edge prior, our method produces complete and sharp edges as desired. Note that more visual results are presented in the supplementary material due to space limitation.
\vspace{-8mm}

\begin{figure}[htbp]
	\centering
	\begin{minipage}[t]{0.49\textwidth}
		\includegraphics[width=5.5cm]{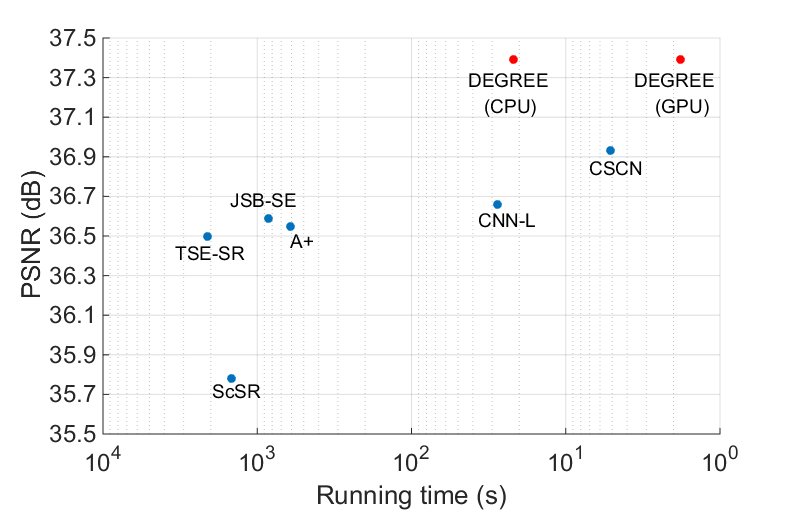}
	\end{minipage}
	\begin{minipage}[t]{0.49\textwidth}
		\includegraphics[width=5.5cm]{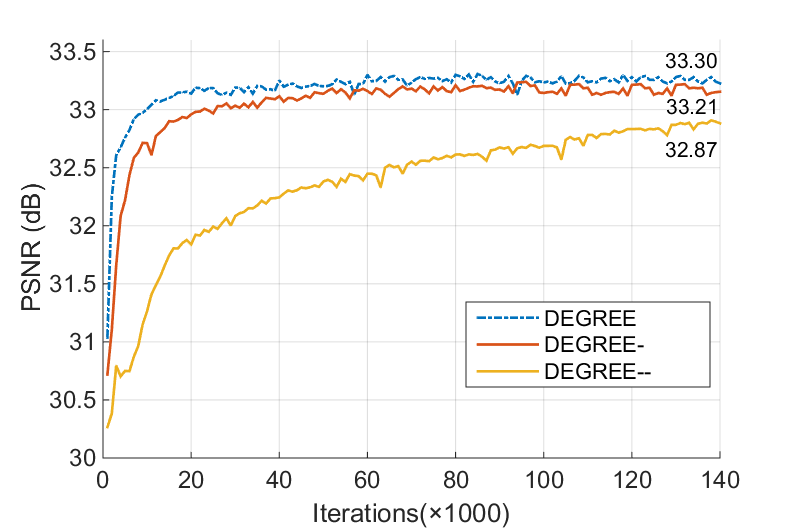}
	\end{minipage}
	
	\begin{minipage}[t]{0.49\textwidth}
		\vspace{-6mm}
		
		\caption{The performance of our method compared with state-of-the-art methods, including the effectiveness and time complexity, in 2$\times$ enlargement on dataset Set5.}
		\label{fig:TVSPerf} 
	\end{minipage} \ 
	\begin{minipage}[t]{0.49\textwidth}
		\vspace{-6mm}
		
		\caption{The comparison of three versions of the proposed method in 3$\times$ enlargement on dataset Set5.
		}
		\label{fig:training} 
	\end{minipage}
	\vspace{-7mm}
\end{figure}
\vspace{-5mm}

%
\vspace{-1mm}

\subsection{Running Time}
\vspace{-1mm}

We report  time cost of our proposed model and compare its efficiency with other methods.
Figure \ref{fig:TVSPerf} plots their running time (in secs.) against performance (in PSNR). All the compared methods are implemented using the public available codes from the authors.
We implement our method using Caffe with its Matlab wrapper. We evaluate the running time of all the algorithms with following machine configuration: Intel X5675 3.07GHz and 24 GB memory. \RRRRM{The GPU version of our method costs 1.81 seconds for performing SR on all the images of Set5, while other methods are significantly slower than ours in orders. The CPU version of our method is comparable to other deep learning-based SR methods, including CSCN and CNN-L.}
\vspace{-5mm}

\subsection{Discussions}
\vspace{-2mm}

We further provide additional investigations on our model in depth, aiming to give more transparent understandings on its effectiveness. 
\vspace{-5mm}

\subsubsection{Ablation Analysis}
We here perform ablation studies to see the individual contribution of each component in our model to the final performance.
 We first observe that, \emph{without} by-pass connections, the training process of our proposed model  could not converge within 40,000 iterations, for various  learning rates (0.1, 0.01 and 0.001). This demonstrates that adding by-pass connections indeed speeds up the convergence rate of  a deep network.
 In the following experiments, we always keep the bypass connections and  evaluate the performance of the following three variants of our model: a vanilla one without edge prior or frequency combination (denoted as DEGREE$--$), the one without frequency combination (denoted as DEGREE$-$) and the full model. Figure~\ref{fig:training} shows their training performance (plotted in curves against number of iterations) and  testing performance (shown in digits) in PSNR on the dataset Set5, for  $\times 3$ enlargement. 
From the results, one can observe that modeling the edge prior boosts the performance significantly and introducing frequency combination further improves the performance.

\vspace{-9mm}

\input{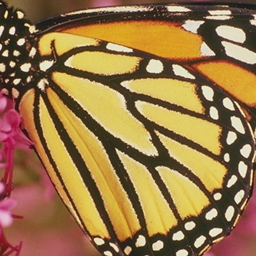}

\subsubsection{Model Size}
We investigate how  the size of the model, including number of layers and  size of channels within each layer, influences the final performance. 
We compare performance of our model with different pairs of (\# layers, \# channels) in Figure~\ref{fig:paraAll}. It can be seen that a large model with more than $(20, 32)\times 10^5$ and $(8, 64)\times 10^5$ parameters (shown as yellow points) is necessary for achieving reasonably good performance. 
The combination of $(20, 8) \times 10^4$ (the purple point) results in a model with the same size of SCN64 (the green point where its dictionary size is equal to 64) and achieves almost the same performance. Further increasing the model size to $(20, 16)\times 10^4$ (the higher purple point) gives a better result than 
SCN128 (with a dictionary size of 64), whose model size is slightly~smaller.
\vspace{-5mm}

\subsubsection{Visualization of Learned \RsixM{Sub-Bands}}
\label{exp:Visualization}
\begin{figure}[htbp]
	\centering
	\begin{minipage}[b]{0.49\textwidth}
		\subfigure{
			\includegraphics[width=5.5cm]{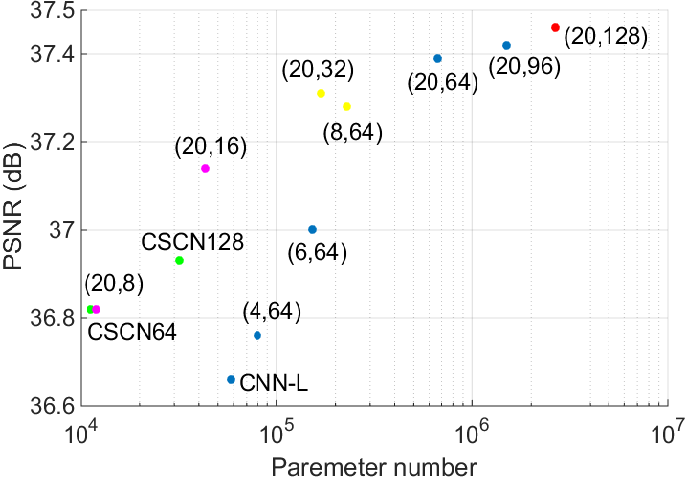}				
		}
	\end{minipage}
	\begin{minipage}[b]{0.50\textwidth}
		\raggedright
		\subfigure[LR]{
			\includegraphics[width=1.7cm]{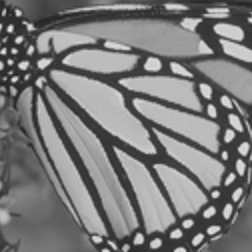}
		}\hspace{-2mm}
		\subfigure[1L]{
			\includegraphics[width=1.7cm]{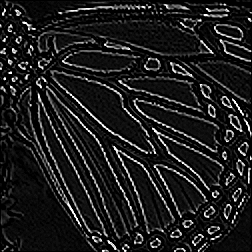}
		}\hspace{-2mm}
		\subfigure[1R]{
			\includegraphics[width=1.7cm]{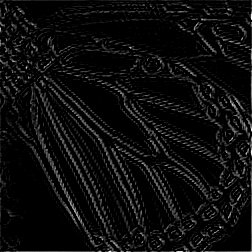}
		}\hspace{-2mm}\\ \vspace{-3.4mm}
		\subfigure[2R]{
			\includegraphics[width=1.7cm]{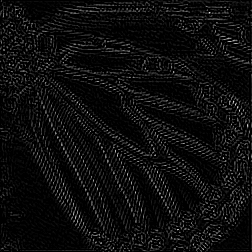}
		}\hspace{-2mm}
		\subfigure[3R]{
			\includegraphics[width=1.7cm]{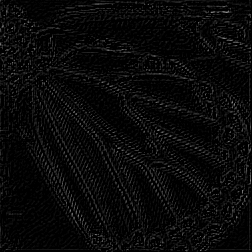}
		}\hspace{-2mm}
		\subfigure[4R]{
			\includegraphics[width=1.7cm]{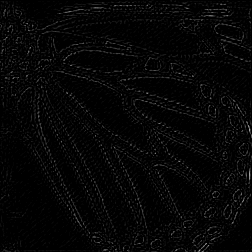}
		}			
	\end{minipage}
	\vspace{-8mm}
	
	\begin{minipage}[t]{0.49\textwidth}
		\caption{PSNR for 2$\times$ SR on Set5 with various parameter numbers, compared with CSCN and CNN.}
		\label{fig:paraAll} 
	\end{minipage}\ \
	\begin{minipage}[t]{0.49\textwidth}
		\caption{The visualization of the \RRM{learned sub-bands} in the recovery on \textit{Butterfly}.
		}
		\label{fig:decomposition} 
	\end{minipage}
	\vspace{-3mm}
	
\end{figure}

We also visualize the learned features from the bottom feature extraction layer (denoted as 1L)  and four recurrent time steps (denoted as $\cdot$R). The results are produced by a network with 10 layers for the $\times 2$ testing case. 
The reconstructed results of \textit{Butterfly} at different layers are shown in Figure~\ref{fig:decomposition}. One can observe that the proposed model captures details at different frequencies, similar to sub-band decomposition. \RRRRM{The 1L layer extracts and enhances the edge features remarkably but brings some artifacts. The 1R layer enhances edges and makes up some false enhancements. In 2R and 3R, the sub-bands contain textures. The 4R layer fixes details. In all, for the whole network, previous layers' sub-bands contain edge features. Later ones include texture features. The sub-band of the last layer models the ``residual signal". \textbf{More visual results are presented in the supplementary material.}}
\vspace{-5mm}

\RRRRRM{
\subsubsection{Application in JPEG Artifacts Reduction}
\noindent It is worth mentioning that the DEGREE network is a general framework in which the prior knowledge is embedded, by properly setting $g(\mathbf{y})$ in  $\mathbf{f}_\text{input}$ and replacing $\mathbf{f}_\text{edge}$ with the feature maps representing specific priors. For example, for JPEG artifacts reduction, we take as input the edge maps of the compressed image and the block map of DCT transformation, \emph{i.e.} $g(\mathbf{y})$, a part of preliminary feature maps. Then we let the network predict $\mathbf{f}_\text{edge}$ consisting of the general edge maps and the edges only overlapped with the block boundary of the high-quality one, which are a part of feature maps of the penultimate layer. The block and feature maps in fact impose the priors about the blockness and edges on the network. \textbf{Results about DEGREE on JPEG artifacts reduction are presented in the supplementary~material.}}
\vspace{-5mm}

%
%

\input{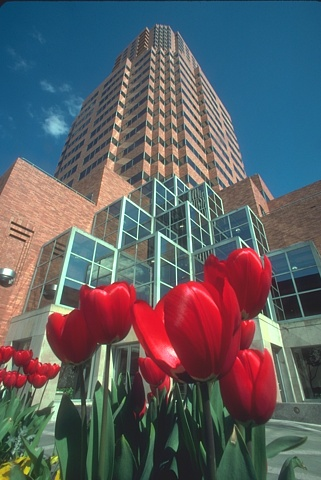}
\input{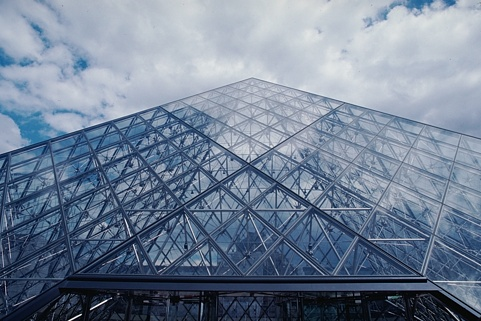}

\section{Conclusions}
\vspace{-2mm}

In this paper, we proposed a deep edge guided recurrent residual network for image SR.  
The edge information is separated out from the image signal to guide the recovery of the HR image. The extracted LR edge maps are used as parts of the input features and the HR edge maps are utilized to constrain the learning of parts of feature maps for image reconstruction. The recurrent residual learning structure with by-pass connections enables the training of deeper networks. Extensive experiments have validated the effectiveness of our method for producing HR images with richer details. Furthermore, this paper presented a general framework for embedding various natural image priors into image processing tasks.

\clearpage

\bibliographystyle{splncs}
\bibliography{egbib}
\end{document}

%% file: butterfly.tex
\def\width{2.6cm}
\def\subwidth{1.3cm}
\def\imagename{butterfly}
\def\folder{.}
\def\scale{4}
\def\imagetitle{butterfly}
\def\startx{50}
\def\starty{0}

\begin{figure}[h]
	\begin{center}
		{
			\subfigure[High-res]{
				\begin{overpic}[width=\width]{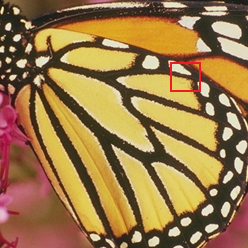}%
					\put(\startx,\starty){\includegraphics[width=\subwidth,clip]{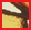}}
				\end{overpic}
			}
			\subfigure[A+]{
				\begin{overpic}[width=\width]{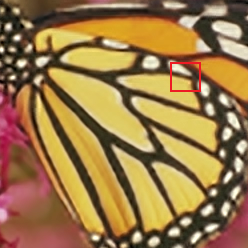}%
					\put(\startx,\starty){\includegraphics[width=\subwidth,clip]{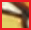}}
				\end{overpic}
			}
			\subfigure[SRCNN]{
				\begin{overpic}[width=\width]{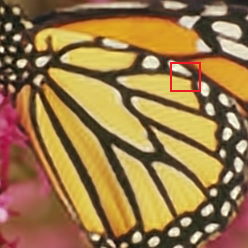}%
					\put(\startx,\starty){\includegraphics[width=\subwidth,clip]{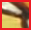}}
				\end{overpic}
			}
			\\\vspace{-3mm}
			\subfigure[JSB-NE]{
				\begin{overpic}[width=\width]{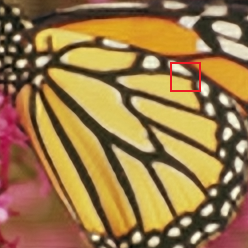}%
					\put(\startx,\starty){\includegraphics[width=\subwidth,clip]{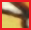}}
				\end{overpic}
			}				
			\subfigure[CSCN]{
				\begin{overpic}[width=\width]{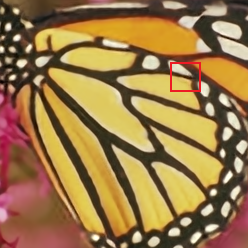}%
					\put(\startx,\starty){\includegraphics[width=\subwidth,clip]{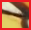}}
				\end{overpic}
			}
			\subfigure[DEGREE]{
				\begin{overpic}[width=\width]{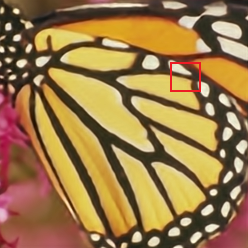}%
					\put(\startx,\starty){\includegraphics[width=\subwidth,clip]{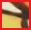}}
				\end{overpic}
			}								
		}
	\end{center}
	\vspace{-8mm}
	
	\caption{
		Visual comparisons between different algorithms for the image \textit{\imagetitle} (\scale$\times$). \RsixM{The DEGREE avoids the artifacts near the corners of the white and yellow plaques.}
	}
	\vspace{-12mm}
	
	\label{fig:butterfly}
\end{figure}

%% file: 86000.tex
\def\width{3cm}
\def\subwidth{1.3cm}
\def\imagename{86000}
\def\folder{.}
\def\scale{3}
\def\imagetitle{86000}
\def\startx{0}
\def\starty{0}

	\begin{figure}[!h]
		\begin{center}
			{
				\subfigure[High-res]{
					\begin{overpic}[width=\width]{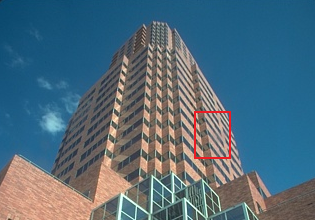}%
						\put(\startx,\starty){\includegraphics[width=\subwidth,clip]{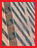}}
					\end{overpic}
				}
				\subfigure[A+]{
					\begin{overpic}[width=\width]{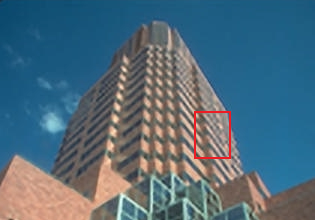}%
						\put(\startx,\starty){\includegraphics[width=\subwidth,clip]{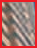}}
					\end{overpic}
				}
				\subfigure[SRCNN]{
					\begin{overpic}[width=\width]{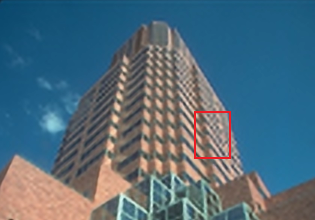}%
						\put(\startx,\starty){\includegraphics[width=\subwidth,clip]{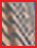}}
					\end{overpic}
				}
				\\\vspace{-3mm}
				\subfigure[\RsixM{JSB-NE}]{
					\begin{overpic}[width=\width]{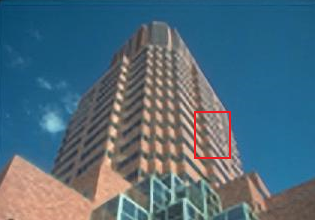}%
						\put(\startx,\starty){\includegraphics[width=\subwidth,clip]{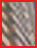}}
					\end{overpic}
				}				
				\subfigure[CSCN]{
					\begin{overpic}[width=\width]{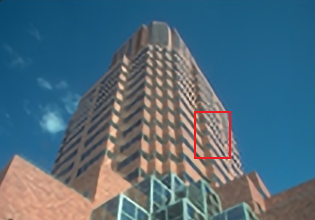}%
						\put(\startx,\starty){\includegraphics[width=\subwidth,clip]{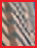}}
					\end{overpic}
				}
				\subfigure[DEGREE]{
					\begin{overpic}[width=\width]{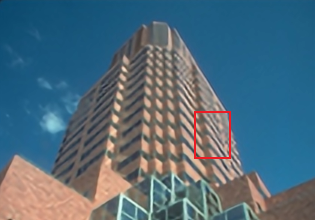}%
						\put(\startx,\starty){\includegraphics[width=\subwidth,clip]{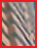}}
					\end{overpic}
				}								
			}
		\end{center}
		\vspace{-7mm}
		
		\caption{
			Visual comparisons between different algorithms for the image \textit{\imagetitle} (\scale$\times$). \RsixM{The DEGREE presents less artifacts around the window boundaries.}
		}
		\vspace{-2mm}
		
		\label{fig:86000}
	\end{figure}

%% file: 223061.tex
\def\width{3cm}
\def\subwidth{1.7cm}
\def\imagename{223061}
\def\folder{.}
\def\scale{3}
\def\imagetitle{223061}
\def\startx{44}
\def\starty{0}

	\begin{figure}[!h]
		\begin{center}
			{
				\subfigure[High-res]{
					\begin{overpic}[width=\width]{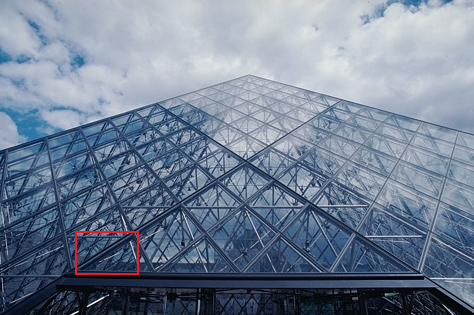}%
						\put(\startx,\starty){\includegraphics[width=\subwidth,clip]{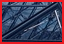}}
					\end{overpic}
				}
				\subfigure[A+]{
					\begin{overpic}[width=\width]{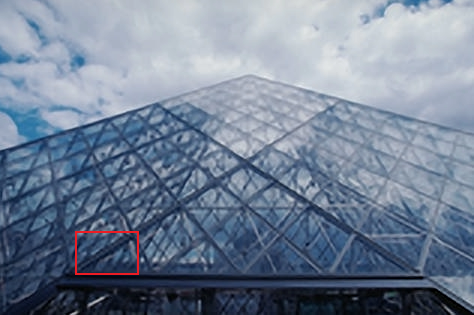}%
						\put(\startx,\starty){\includegraphics[width=\subwidth,clip]{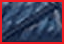}}
					\end{overpic}
				}
				\subfigure[SRCNN]{
					\begin{overpic}[width=\width]{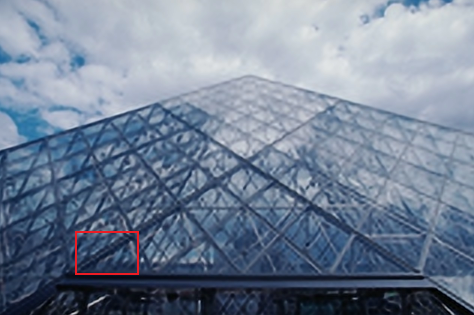}%
						\put(\startx,\starty){\includegraphics[width=\subwidth,clip]{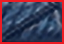}}
					\end{overpic}
				}
				\\\vspace{-3mm}
				\subfigure[TSE-SR]{
					\begin{overpic}[width=\width]{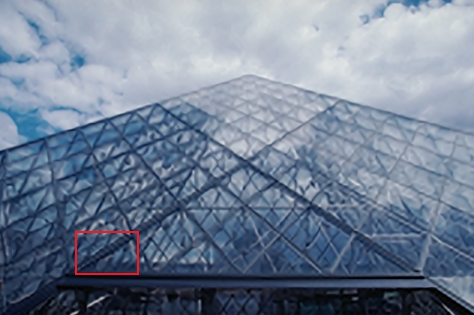}%
						\put(\startx,\starty){\includegraphics[width=\subwidth,clip]{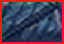}}
					\end{overpic}
				}				
				\subfigure[CSCN]{
					\begin{overpic}[width=\width]{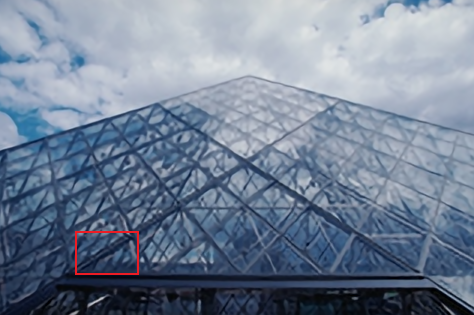}%
						\put(\startx,\starty){\includegraphics[width=\subwidth,clip]{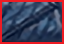}}
					\end{overpic}
				}
				\subfigure[DEGREE]{
					\begin{overpic}[width=\width]{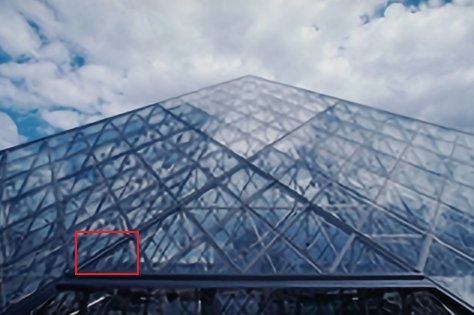}%
						\put(\startx,\starty){\includegraphics[width=\subwidth,clip]{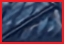}}
					\end{overpic}
				}								
			}
		\end{center}
		\vspace{-8mm}
		
		\caption{
			Visual comparisons between different algorithms for the image \textit{\imagetitle} (\scale$\times$). \RsixM{The DEGREE produces more complete and sharper edges.}
		}
		\vspace{-6mm}
		
		\label{fig:223061}
	\end{figure}